\documentclass[runningheads]{llncs}
\usepackage{cite}
\usepackage{graphicx}
\usepackage{amsmath}
\usepackage{verbatim}
\usepackage{multirow}
\usepackage{amssymb}
\usepackage{caption}
\usepackage{floatrow}
\usepackage{subfigure}
\usepackage{multirow}
\usepackage{comment}
\usepackage{siunitx}
\usepackage{booktabs}
\usepackage{color}
\usepackage{hyperref}
\hypersetup{
    colorlinks=true,
    linkcolor=blue,
    filecolor=magenta,      
    urlcolor=red,
}

\usepackage{bm}
\usepackage[utf8]{inputenc}

\newcommand*{\op}[1]{\operatorname{#1}}

\newfloatcommand{capbtabbox}{table}[][\FBwidth]
\begin{document}
\title{TransUNet: Transformers Make Strong Encoders for Medical Image Segmentation}
\author{Jieneng Chen\textsuperscript{1}, Yongyi Lu\textsuperscript{1}, Qihang Yu\textsuperscript{1},
Xiangde Luo\textsuperscript{2}, \\Ehsan Adeli\textsuperscript{3}, Yan Wang\textsuperscript{4}, Le Lu\textsuperscript{5}, Alan L. Yuille\textsuperscript{1}, and Yuyin Zhou\textsuperscript{3}}

\authorrunning{J. Chen et al.}

\institute{
\textsuperscript{1}Johns Hopkins University\\
\textsuperscript{2}University of Electronic Science and Technology of China\\
\textsuperscript{3}Stanford University\\
\textsuperscript{4} East China Normal University \\
\textsuperscript{5}PAII Inc.\\ }
\maketitle
\begin{abstract}
Medical image segmentation is an essential prerequisite for developing healthcare systems, especially for disease diagnosis and treatment planning. 
On various medical image segmentation tasks, the u-shaped architecture, also known as U-Net, has become the de-facto standard and achieved tremendous success.
However, due to the intrinsic locality of convolution operations, U-Net generally demonstrates limitations in explicitly modeling long-range dependency.
Transformers, designed for sequence-to-sequence prediction, have emerged as alternative architectures with innate global self-attention mechanisms, but can result in limited localization abilities due to insufficient low-level details.
In this paper, we propose~\textbf{TransUNet}, which merits both Transformers and U-Net, as a strong alternative for medical image segmentation.
On one hand, the Transformer encodes tokenized image patches from a convolution neural network (CNN) feature map as the input sequence for extracting global contexts. On the other hand, the decoder upsamples the encoded features which are then combined with the high-resolution CNN feature maps to enable precise localization.

We argue that Transformers can serve as strong encoders for medical image segmentation tasks, with the combination of U-Net to enhance finer details by recovering localized spatial information. TransUNet achieves superior performances to various competing methods on different medical applications including multi-organ segmentation and cardiac segmentation. Code and models are available at~\url{https://github.com/Beckschen/TransUNet}.

\end{abstract}


%
%
%
\section{Introduction}
Convolutional neural networks (CNNs), especially fully convolutional networks (FCNs)~\cite{long2015fully}, have become dominant in medical image segmentation. Among different variants, U-Net~\cite{ronneberger2015u}, which consists of a symmetric encoder-decoder network with skip-connections to enhance detail retention, has become the de-facto choice.
Based on this line of approach, tremendous success has been achieved in a wide range of medical applications such as cardiac segmentation from magnetic resonance (MR)~\cite{yu2017automatic}, organ segmentation from computed tomography (CT) \cite{zhou2017fixed,li2018h,yu2018recurrent} and polyp segmentation~\cite{zhou2018unet++} from colonoscopy videos.

In spite of their exceptional representational power,  CNN-based approaches generally exhibit limitations for modeling explicit long-range relation, due to the intrinsic locality of convolution operations.
Therefore, these architectures generally yield weak performances especially for target structures that show large inter-patient variation in terms of texture, shape and size.
To overcome this limitation, existing studies propose to establish self-attention mechanisms based on CNN features~\cite{wang2018non,schlemper2019attention}.
On the other hand, Transformers, designed for sequence-to-sequence prediction, have emerged as alternative architectures which employ dispense convolution operators entirely and solely rely on attention mechanisms instead~\cite{vaswani2017attention}. 
Unlike prior CNN-based methods, Transformers are not only powerful at modeling global contexts but also demonstrate superior transferability for downstream tasks under large-scale pre-training.
The success has been widely witnessed in the field of machine translation and natural language processing (NLP)~\cite{vaswani2017attention,devlin2018bert}.
More recently, attempts have also matched or even exceeded state-of-the-art performances for various image recognition tasks~\cite{dosovitskiy2020image,zheng2020rethinking}. 

In this paper, we present the first study which explores the potential of transformers in the context of medical image segmentation. 
However, interestingly, we found that a naive usage ({\em i.e.}, use a transformer for encoding the tokenized image patches, and then directly upsamples the hidden feature representations into a dense output of full resolution) cannot produce a satisfactory result.
 
This is due to that Transformers treat the input as 1D sequences and exclusively focus on modeling the global context at all stages, therefore result in low-resolution features which lack detailed localization information.
And this information cannot be effectively recovered by direct upsampling to the full resolution, therefore leads to a coarse segmentation outcome.
On the other hand, CNN architectures (\emph{e.g.}, U-Net~\cite{ronneberger2015u}) provide an avenue for extracting low-level visual cues which can well remedy such fine spatial details.

To this end, we propose \textbf{TransUNet}, the first medical image segmentation framework, which establishes self-attention mechanisms from the perspective of sequence-to-sequence prediction.
To compensate for the loss of feature resolution brought by Transformers, TransUNet employs a hybrid CNN-Transformer architecture to leverage both detailed high-resolution spatial information from CNN features and the global context encoded by Transformers.
Inspired by the u-shaped architectural design, the self-attentive feature encoded by Transformers is then upsampled to be combined with different high-resolution CNN features skipped from the encoding path, for enabling precise localization. 
We show that such a design allows our framework to preserve the advantages of Transformers and 
also benefit medical image segmentation. 
Empirical results suggest that our Transformer-based architecture presents a better way to leverage self-attention compared with previous CNN-based self-attention methods.
Additionally, we observe that more intensive incorporation of low-level features generally leads to a better segmentation accuracy. 
Extensive experiments demonstrate the superiority of our method against other competing methods on various medical image segmentation tasks.

\section{Related Works}
\noindent\textbf{Combining CNNs with self-attention mechanisms.}
Various studies have attempted to integrate self-attention mechanisms into CNNs by modeling global interactions of all pixels based on the feature maps. For instance, Wang \emph{et al.} designed a non-local operator, which can be plugged into multiple intermediate convolution layers~\cite{wang2018non}. Built upon the encoder-decoder u-shaped architecture, Schlemper~\emph{et al.}~\cite{schlemper2019attention} proposed additive attention gate modules which are integrated into the skip-connections.
Different from these approaches, we employ Transformers for embedding global self-attention in our method.

\vspace{1ex}\noindent\textbf{Transformers.}
Transformers were first proposed by \cite{vaswani2017attention} for machine translation and established state-of-the-arts in many NLP tasks.
To make Transformers also applicable for computer vision tasks, several modifications have been made.
For instance, Parmar \emph{et al.}~\cite{parmar2018image} applied the self-attention only in local neighborhoods for each query pixel instead of globally.
Child \emph{et al.}~\cite{child2019generating} proposed Sparse Transformers, which employ scalable approximations to global self-attention.
Recently, Vision Transformer (ViT)~\cite{dosovitskiy2020image} achieved state-of-the-art on ImageNet classification by directly applying Transformers with global self-attention to full-sized images. 
To the best of our knowledge, the proposed TransUNet is the first Transformer-based medical image segmentation framework, which builds upon the highly successful ViT.

\section{Method}
Given an image $\bm{\mathrm{x}} \in \mathbb{R}^{H \times W \times C}$ with an spatial resolution of $H \times W$ and $C$ number of channels. Our goal is to predict the corresponding pixel-wise labelmap with size $H \times W$. The most common way is to directly train a CNN (\emph{e.g.}, U-Net) to first encode images into high-level feature representations, which are then decoded back to the full spatial resolution.
Unlike existing approaches, our method introduces self-attention mechanisms into the encoder design via the usage of Transformers.
We will first introduce how to directly apply transformer for encoding feature representations from decomposed image patches in Section~\ref{sec:transformer_encoder}.
Then, the overall framework of TransUNet will be elaborated in Section~\ref{sec:transunet}.

\subsection{Transformer as Encoder}
\label{sec:transformer_encoder}
\noindent\textbf{Image Sequentialization.} 
Following~\cite{dosovitskiy2020image}, we first perform tokenization by reshaping the input $\bm{\mathrm{x}}$ into a sequence of flattened 2D patches \{$\bm{\mathrm{x}}^i_p \in \mathbb{R}^{P^2 \cdot C}|i=1,..,N\}$, where each patch is of size $P \times P$ and $N=\frac{HW}{P^2}$ is the number of image patches (\emph{i.e.}, the input sequence length). 

\vspace{1ex}\noindent\textbf{Patch Embedding.} We map the vectorized patches $\bm{\mathrm{x}}_p$ into a latent $D$-dimensional embedding space using a trainable linear projection.
To encode the patch spatial information, we learn specific position embeddings which are added to the patch embeddings to retain positional information as follows:

\begin{align}
    \bm{\mathrm{z}}_0 &= [\bm{\mathrm{x}}^1_p \bm{\mathrm{E}}; \, \bm{\mathrm{x}}^2_p \bm{\mathrm{E}}; \cdots; \, \bm{\mathrm{x}}^{N}_p \bm{\mathrm{E}} ] + \bm{\mathrm{E}}_{pos}, \label{eq:embedding} 
\end{align}

\noindent where $\bm{\mathrm{E}} \in \mathbb{R}^{(P^2 \cdot C) \times D}$ is the patch embedding projection, and $\bm{\mathrm{E}}_{pos}  \in \mathbb{R}^{N \times D}$ denotes the position embedding.

\begin{figure*}[t!]
    \centering
    \includegraphics[width=\textwidth]{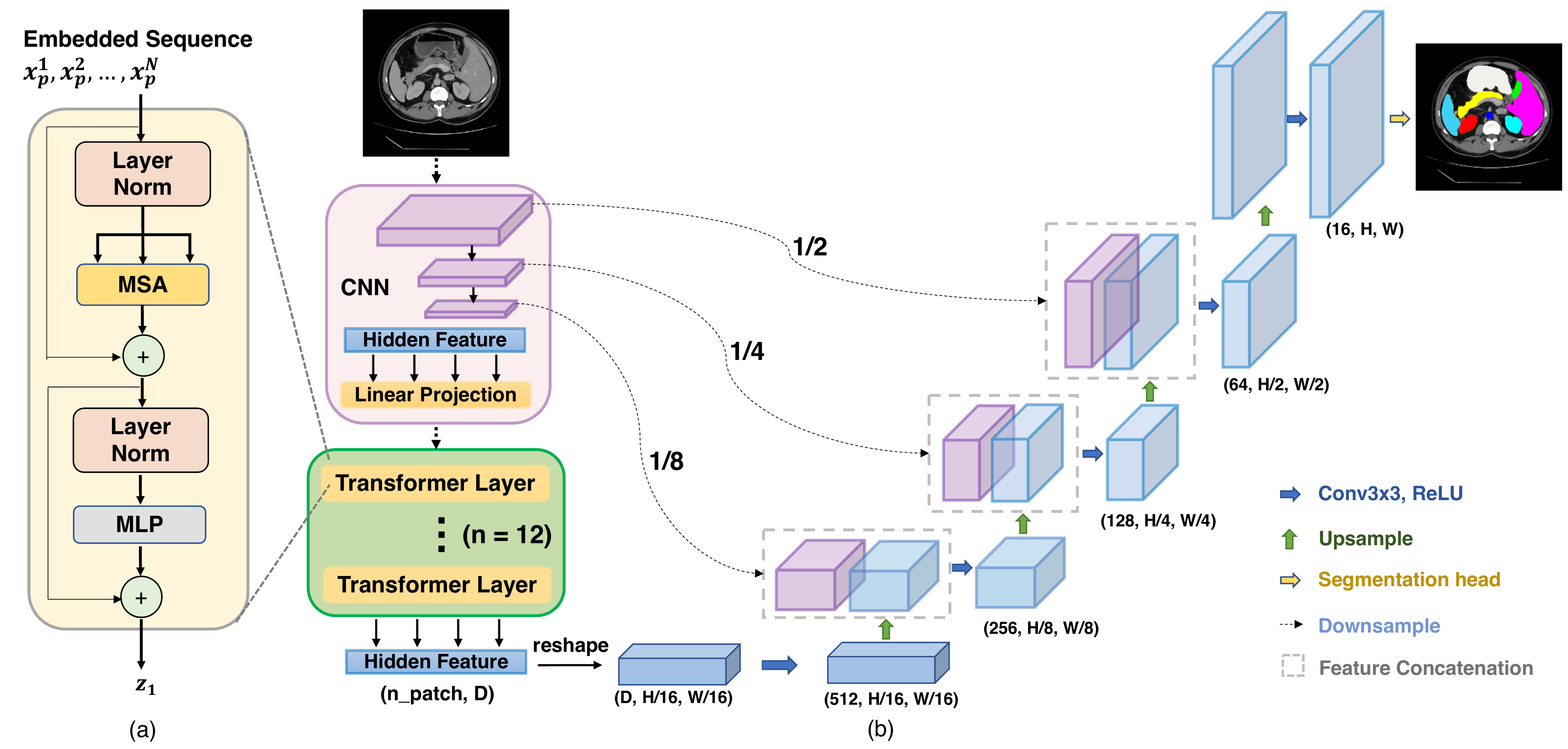}
    \caption{Overview of the framework. (a) schematic of the Transformer layer; (b) architecture of the proposed TransUNet.
    }
    \label{fig:framework}
\end{figure*}

The Transformer encoder consists of $L$ layers of Multihead Self-Attention (MSA) and Multi-Layer Perceptron (MLP) blocks (Eq.~\eqref{eq:msa_apply}\eqref{eq:mlp_apply}). Therefore the output of the $\ell$-th layer can be written as follows:
\begin{align}
    \bm{\mathrm{z}}^\prime_\ell &= \op{MSA}(\op{LN}(\bm{\mathrm{z}}_{\ell-1})) + \bm{\mathrm{z}}_{\ell-1}, &&  \label{eq:msa_apply} \\
    \bm{\mathrm{z}}_\ell &= \op{MLP}(\op{LN}(\bm{\mathrm{z}}^\prime_{\ell})) + \bm{\mathrm{z}}^\prime_{\ell},   \label{eq:mlp_apply} 
\end{align}
where $\op{LN}(\cdot)$ denotes the layer normalization operator and $\bm{\mathrm{z}}_L$ is the encoded image representation. The structure of a Transformer layer is illustrated in Figure~\ref{fig:framework}(a).

\subsection{TransUNet}
\label{sec:transunet}
For segmentation purposes, an intuitive solution is to simply upsample the encoded feature representation $\bm{\mathrm{z}}_L \in \mathbb{R}^{\frac{HW}{P^2} \times D}$ to the full resolution for predicting the dense output.
Here to recover the spatial order, the size of the encoded feature should first be reshaped from $\frac{HW}{P^2}$  to $\frac{H}{P} \times \frac{W}{P}$.
We use a  $1 \times 1$ convolution to reduce channel size of the reshaped feature to number of class, and then the feature map is directly bilinearly upsampled to the full resolution $H \times W$ for predicting the final segmentation outcome. In later comparisons in Section \ref{sec:comparison}, we denote this naive upsampling baseline as ``None'' in the decoder design. 

Although combining a Transformer with naive upsampling already yields a reasonable performance, as mentioned above, this strategy is not the optimal usage of Transformers in segmentation since $\frac{H}{P} \times \frac{W}{P}$ is usually much smaller than the original image resolution $H \times W$, therefore inevitably results in a loss of low-level details (\emph{e.g.}, shape and boundary of the organ).
Therefore, to compensate for such information loss, TransUNet employs a hybrid CNN-Transformer architecture as the encoder as well as a cascaded upsampler to enable precise localization. The overview of the proposed TransUNet is depicted in Figure~\ref{fig:framework}.

\vspace{1ex}\noindent\textbf{CNN-Transformer Hybrid as Encoder.}
Rather than using the pure Transformer as the encoder (Section~\ref{sec:transformer_encoder}), TransUNet employs a  CNN-Transformer hybrid model where CNN is first used as a feature extractor to generate a feature map for the input. Patch embedding is applied to $1\times1$ patches extracted from the CNN feature map instead of from raw images. 

We choose this design since 1) it allows us to leverage the intermediate high-resolution CNN feature maps in the decoding path; and 2) we find that the hybrid CNN-Transformer encoder performs better than simply using a pure Transformer as the encoder.

\vspace{1ex}\noindent\textbf{Cascaded Upsampler.}
We introduce a cascaded upsampler (CUP), which consists of multiple upsampling steps to decode the hidden feature for outputting the final segmentation mask.
After reshaping the sequence of hidden feature $\bm{\mathrm{z}}_L \in \mathbb{R}^{\frac{HW}{P^2} \times D}$ to  the shape of $\frac{H}{P} \times \frac{W}{P} \times D$, we instantiate CUP by cascading multiple upsampling blocks for reaching the full resolution from $\frac{H}{P} \times \frac{W}{P}$  to $H \times W$, where each block consists of a $2 \times$ upsampling operator, a 3$\times$3 convolution layer, and a ReLU layer successively.

We can see that CUP together with the hybrid encoder form a u-shaped architecture which enables feature aggregation at different resolution levels via skip-connections.
The detailed architecture of CUP as well as the intermediate skip-connections can be found in Figure~\ref{fig:framework}(b).

\section{Experiments and Discussion}
\subsection{Dataset and Evaluation}
Synapse multi-organ segmentation dataset\footnote{\url{https://www.synapse.org/\#!Synapse:syn3193805/wiki/217789}}.
We use the 30 abdominal CT scans in the MICCAI 2015 Multi-Atlas Abdomen Labeling Challenge, with $3779$ axial contrast-enhanced abdominal clinical CT images in total. 

Each CT volume consists of $85\sim198$ slices of $512\times512$ pixels, with a voxel spatial resolution of $([0.54\sim0.54]\times[0.98\sim0.98]\times[2.5\sim5.0])\textup{mm}^3$. Following~\cite{fu2020domain}, we report the average DSC and average Hausdorff Distance (HD) on 8 abdominal organs (aorta, gallbladder, spleen, left kidney, right kidney, liver, pancreas, spleen, stomach with a random split of 18 training cases (2212 axial slices) and 12 cases for validation.

\noindent\textbf{Automated cardiac diagnosis challenge\footnote{\url{https://www.creatis.insa-lyon.fr/Challenge/acdc/}}.}   The ACDC challenge collects exams from different patients acquired from MRI scanners. Cine MR images were acquired in breath hold, and a series of short-axis slices cover the heart from the base to the apex of the left ventricle, with a slice thickness of 5 to 8 mm. The short-axis in-plane spatial resolution goes from 0.83 to 1.75 mm$^2$/pixel.

Each patient scan is manually annotated with ground truth for left ventricle (LV), right ventricle (RV) and myocardium (MYO). We report the average DSC with a random split of 70 training cases (1930 axial slices), 10 cases for validation and 20 for testing.

\vspace{-1em}
\begin{table}[]
\footnotesize
\renewcommand\tabcolsep{1.0pt}
\resizebox{\textwidth}{!}
{
\begin{tabular}{cccccccccccc}
\hline
\multicolumn{2}{c}{Framework} & \multicolumn{2}{c}{\textbf{Average}} & \multirow{2}{*}{Aorta} & \multirow{2}{*}{Gallbladder} & \multirow{2}{*}{Kidney (L)} & \multirow{2}{*}{Kidney (R)} & \multirow{2}{*}{Liver} & \multirow{2}{*}{Pancreas} & \multirow{2}{*}{Spleen} & \multirow{2}{*}{Stomach} \\
\cmidrule(lr){1-2} \cmidrule(lr){3-4} 
Encoder & Decoder & DSC~$\uparrow$ & HD~$\downarrow$         
&          &         &             &            &             &           &             &           \\ 
\hline
\multicolumn{2}{c}{V-Net~\cite{milletari2016v}}     & 68.81        & -            & 75.34          & 51.87              & 77.10             & 80.75             & 87.84           & 40.05            & 80.56          & 56.98           \\
\multicolumn{2}{c}{DARR~\cite{fu2020domain}}      & 69.77        & -            & 74.74           & 53.77             & 72.31            & 73.24             & 94.08           & 54.18            & 89.90           & 45.96            \\
R50        & U-Net~\cite{ronneberger2015u}             & 74.68        & 36.87        & 84.18      & 62.84          & 79.19         & 71.29          & 93.35        & 48.23       & 84.41       & 73.92        \\
R50        & \scriptsize{AttnUNet}~\cite{schlemper2019attention} & 75.57   & 36.97        & 55.92      & 63.91          & 79.20         & 72.71          & 93.56       & 49.37        & 87.19       & 74.95        \\ \hline 
ViT~\cite{dosovitskiy2020image}    & None                 & 61.50        & 39.61        & 44.38      & 39.59          & 67.46         & 62.94          & 89.21       & 43.14       & 75.45       & 69.78        \\
ViT~\cite{dosovitskiy2020image}        & CUP              & 67.86        & 36.11        & 70.19      & 45.10          & 74.70         & 67.40         & 91.32       & 42.00        & 81.75       & 70.44       \\
R50-ViT~\cite{dosovitskiy2020image}    & CUP              & 71.29        & 32.87        & 73.73      & 55.13          & 75.80         & 72.20         & 91.51       & 45.99        & 81.99      & 73.95        \\
\multicolumn{2}{c}{\textbf{TransUNet}} & \textbf{77.48}        & \textbf{31.69}        & 87.23        & 63.13          & 81.87          & 77.02          & 94.08      & 55.86         & 85.08      & 75.62        \\ 
\hline
\end{tabular}
\vspace{-0.6em}
}
\caption{Comparison on the Synapse multi-organ CT dataset (average dice score \% and average hausdorff distance in mm, and dice score \% for each organ).
}
\label{tab:synapse}
\end{table}

\vspace{-2.5em}
\subsection{Implementation Details}
For all experiments, we apply simple data augmentations, e.g., random rotation and flipping. 

For pure Transformer-based encoder, we simply adopt ViT~\cite{dosovitskiy2020image} with 12 Transformer layers.
For the hybrid encoder design, we combine ResNet-50~\cite{he2016deep} and ViT, denoted as ``R50-ViT'', throught this paper.
All Transformer backbones (\emph{i.e.}, ViT) and ResNet-50 (denoted as ``R-50'') were pretrained on ImageNet~\cite{deng2009imagenet}. 
The input resolution and patch size $P$ are set as 224$\times$224 and 16, unless otherwise specified.
Therefore, we need to cascade four $2 \times$ upsampling blocks consecutively in CUP to reach the full resolution.
And for 
Models are trained with SGD optimizer with learning rate 0.01, momentum 0.9 and weight decay 1e-4.
The default batch size is 24 and the default number of training iterations are 20k for ACDC dataset and 14k for Synapse dataset respectively. All experiments are conducted using a single Nvidia RTX2080Ti GPU.

Following~\cite{zhou2017fixed,yu2018recurrent}, all 3D volumes are inferenced in a slice-by-slice fashion and the predicted 2D slices are stacked together to reconstruct the 3D prediction for evaluation.

\subsection{Comparison with State-of-the-arts}
\label{sec:comparison}

We conduct main experiments on Synapse multi-organ segmentation dataset by comparing our TransUNet with four previous state-of-the-arts: 1) V-Net~\cite{milletari2016v}; 2) DARR~\cite{fu2020domain}; 3) U-Net~\cite{ronneberger2015u} and 4) AttnUNet~\cite{schlemper2019attention}. 

To demonstrate the effectiveness of our CUP decoder, we use ViT~\cite{dosovitskiy2020image} as the encoder, and compare results using naive upsampling (``None'') and CUP as the decoder, respectively;
To demonstrate the effectiveness of our hybrid encoder design, we use CUP as the decoder, and compare results using ViT and R50-ViT as the encoder, respectively.
In order to make the comparison with the ViT-hybrid baseline (R50-ViT-CUP) and our TransUNet to be fair, we also replace the original encoder of U-Net \cite{ronneberger2015u} and AttnUNet \cite{oktay2018attention} with ImageNet pretrained ResNet-50.
The results in terms of DSC and mean hausdorff distance (in mm) are reported in Table \ref{tab:synapse}.

Firstly, we can see that compared with ViT-None, ViT-CUP observes an improvement of $6.36\%$ and $3.50$ mm in terms of average DSC and Hausdorff distance respectively. This improvement suggests that our CUP design presents a better decoding strategy than direct upsampling. 
Similarly, compared with ViT-CUP, R50-ViT-CUP also suggests an additional improvement of $3.43\%$ in DSC and $3.24$ mm in Hausdorff distance, which demonstrates the effectiveness of our hybrid encoder.
Built upon R50-ViT-CUP, our TransUNet which is also equipped with skip-connections,  achieves the best result among different variants of Transformer-based models.

Secondly, Table \ref{tab:synapse} also shows that the proposed TransUNet has significant improvements over prior arts, \emph{e.g.}, performance gains range from 1.91\% to 8.67\% considering average DSC. 
In particular, directly applying Transformers for multi-organ segmentation yields reasonable results (67.86\% DSC for ViT-CUP), but cannot match the performance of U-Net or attnUNet. 
This is due to that Transformers can well capture high-level semantics which are favorable for classification task but lack of low-level cues for segmenting the fine shape of medical images. 
On the other hand, combining Transformers with CNN, \emph{i.e.}, R50-ViT-CUP, outperforms V-Net and DARR but still yield inferior results than pure CNN-based R50-U-Net and R50-AttnUNet. Finally, when combined with the U-Net structure via skip-connections, the proposed TransUNet sets a new state-of-the-art, outperforming R50-ViT-CUP and previous best R50-AttnUNet by 6.19\% and 1.91\% respectively, showing the strong ability of TransUNet to learn both high-level semantic features as well as low-level details, which is crucial in medical image segmentation.
A similar trend can be also witnessed for the average Hausdorff distance, which further demonstrates the advantages of our TransUNet over these CNN-based approaches.

\subsection{Analytical Study}
\label{sec:analytical}
To thoroughly evaluate the proposed TransUNet framework and validate the performance under different settings, a variety of ablation studies were performed, including: 1) the number of skip-connections; 2) input resolution; 3) sequence length and patch size and 4) model scaling.

\begin{figure*}
    \centering
    \includegraphics[width=0.85\textwidth]{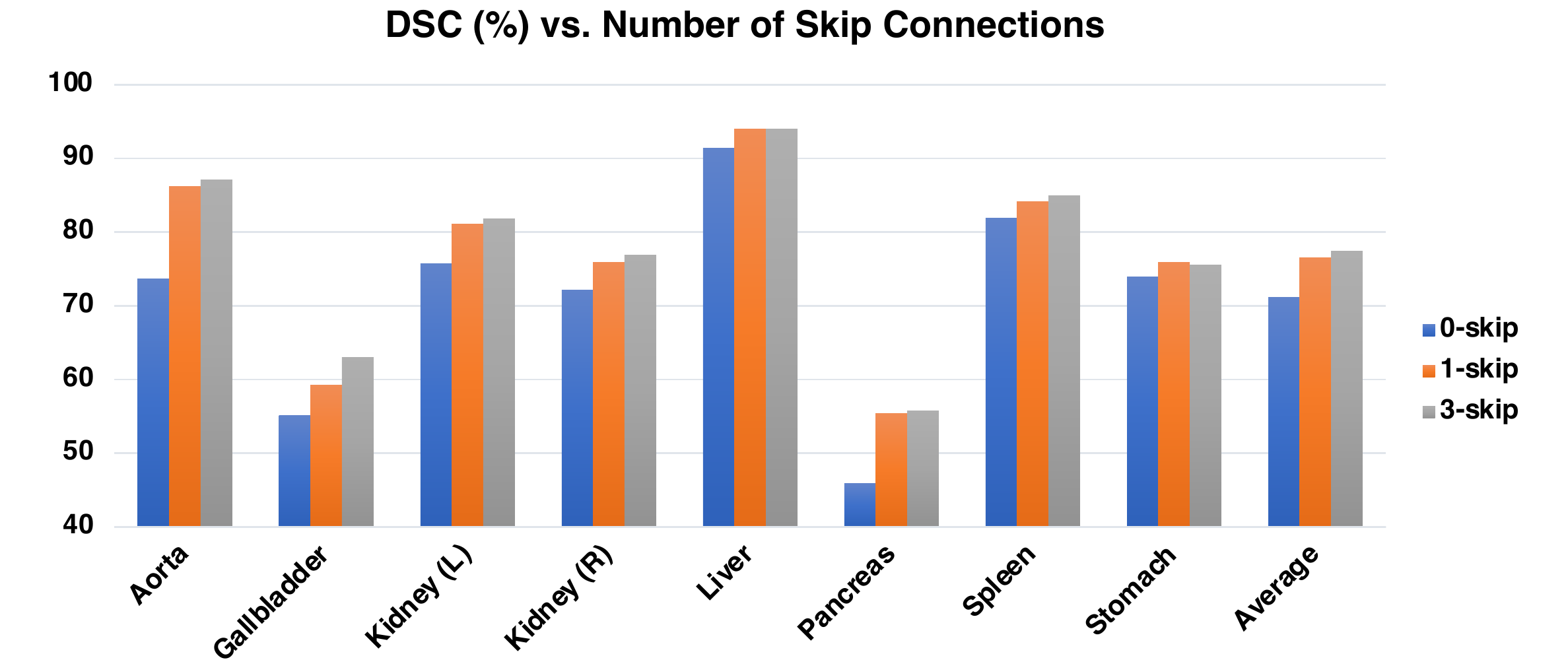}
    \caption{Ablation study on the number of skip-connections in TransUNet.}
    \label{fig:abl:skip}
\end{figure*}

\vspace{1ex}\noindent\textbf{The Number of Skip-connections.}
As discussed above, integrating U-Net-like skip-connections help enhance finer segmentation details by recovering low-level spatial information. The goal of this ablation is to test the impact of adding different numbers of skip-connections in TransUNet. 
By varying the number of skip-connections to be 0 (R50-ViT-CUP)/1/3, the segmentation performance in average DSC on all 8 testing organs are summarized in Figure~\ref{fig:abl:skip}.
Note that in the ``1-skip'' setting, we add the skip-connection only at the 1/4 resolution scale.
We can see that adding more skip-connections generally leads to a better segmentation performance. 
The best average DSC and HD are achieved by inserting skip-connections to all three intermediate upsampling steps of CUP except the output layer, \emph{i.e.}, at 1/2, 1/4, and 1/8 resolution scales (illustrated in Figure \ref{fig:framework}). 
Thus, we adopt this configuration for our TransUNet. 
It is also worth mentioning that the performance gain of smaller organs (\emph{i.e.}, aorta, gallbladder, kidneys, pancreas) is more evident than that of larger organs (\emph{i.e.}, liver, spleen, stomach).
These results reinforce our initial intuition of integrating U-Net-like skip-connections into the Transformer design to enable learning precise low-level details.

As an interesting study, we apply additive Transformers in the skip-connections, similar to~\cite{schlemper2019attention}, and find this new type of skip-connection can even further the segmentation performance. 
Due to the GPU memory constraint, we employ a light Transformer in the 1/8 resolution scale skip-connection while keeping the other two skip-connections unchanged. As a result, this simple alteration leads to a performance boost of 1.4 \% DSC.

\vspace{1ex}\noindent\textbf{On the Influence of Input Resolution.}
The default input resolution for TransUNet is 224$\times$224. Here, we also provide results of training TransUNet on a high-resolution 512$\times$512, as shown in Table \ref{tab:resolution}. When using 512$\times$512 as input, we keep the same patch size (\emph{i.e.}, 16), which results in an approximate 5$\times$ larger sequence length for the Transformer. As \cite{dosovitskiy2020image} indicated, increasing the effective sequence length shows robust improvements. For TransUNet, changing the resolution scale from 224$\times$224 to 512$\times$512 results in 6.88\% improvement in average DSC, at the expense of a much larger computational cost. 
Therefore, considering the computation cost, all experimental comparisons in this paper are conducted with a default resolution of $224 \times 224$ to demonstrate the effectiveness of TransUNet.

\vspace{-1em}
\begin{table}[]
\footnotesize
\resizebox{\textwidth}{!}
{
\begin{tabular}{c|c|ccccccccc}
\hline
Resolution         & Average DSC    & Aorta & Gallbladder & Kidney (L)  & Kidney (R) & Liver & Pancreas & Spleen & Stomach \\ \hline
224   & \textbf{77.48} & 87.23 & 63.13 & 81.87      & 77.02      & 94.08 & 55.86    & 85.08  & 75.62   \\ \hline

512 & \textbf{84.36} & 90.68 & 71.99       & 86.04      & 83.71      & 95.54 & 73.96    & 88.80  & 84.20   \\ \hline
\end{tabular}
}
\caption{Ablation study on the influence of input resolution.}
\label{tab:resolution}
\end{table}

\vspace{-1.5em}
\noindent\textbf{On the Influence of Patch Size/Sequence Length.}

We also investigate the influence of patch size on TransUNet. The results are summarized in Table \ref{tab:ps}. It is
observed that a higher segmentation performance is usually obtained with smaller patch size. 
Note that the Transformer's sequence length is inversely proportional to the square of the patch size (\emph{e.g.}, patch size 16 corresponds to a sequence length of 196 while patch size 32 has a shorter sequence length of 49), therefore decreasing the patch size (or increasing the effective sequence length) shows robust improvements, as the Transformer encodes more complex dependencies between each element for longer input sequences. Following the setting in ViT~\cite{dosovitskiy2020image}, we use 16$\times$16 as the default patch size throughout this paper.

\vspace{-1em}
\begin{table*}[]
\centering
\footnotesize
\resizebox{\textwidth}{!}
{
\begin{tabular}{c|c|c|cccccccc}
\hline
Patch size   & Seq\_length       & Average DSC    & Aorta & Gallbladder & Kidney (L) & Kidney (R) & Liver & Pancreas & Spleen & Stomach \\ \hline
32 & 49   & 76.99 & 86.66 & 63.06       & 81.61      & 79.18      & 94.21 & 51.66    & 85.38  & 74.17   \\ 
16  & 196 & 77.48 & 87.23 & 63.13       & 81.87      & 77.02      & 94.08 & 55.86    & 85.08  & 75.62   \\
8  & 784 & \textbf{77.83} & 86.92 & 58.31       & 81.51      & 76.40      & 93.81 & 58.09    & 87.92  & 79.68   \\ \hline
\end{tabular}
} 
\caption{Ablation study on the patch size and the sequence length.
\vspace{-0.5em}}
\label{tab:ps}
\end{table*}

\vspace{-1.5em}
\noindent\textbf{Model Scaling.}
Last but not least, we provide ablation study on different model sizes of TransUNet. 
In particular, we investigate two different TransUNet configurations, the ``Base'' and ``Large'' models. For the ``base'' model, the hidden size $D$, number of layers, MLP size, and number of heads are set to be 12, 768, 3072, and 12, respectively while those hyperparamters for ``large'' model are 24, 1024, 4096, and 16. From Table \ref{tab:scale} we conclude that larger model results in a better performance. Considering the computation cost, we adopt ``Base'' model for all the experiments.

\vspace{-1em}
\begin{table*}[]
\footnotesize
\centering
\resizebox{\textwidth}{!}
{
\begin{tabular}{c|c|cccccccc}
\hline
Model scale          & Average DSC    & Aorta & Gallbladder & Kidney (L) & Kidney (R) & Liver & Pancreas & Spleen & Stomach \\ \hline
Base   & 77.48 & 87.23 & 63.13       & 81.87      & 77.02      & 94.08 & 55.86    & 85.08  & 75.62   \\
Large & 78.52 & 87.42 & 63.92       & 82.17      & 80.19      & 94.47 & 57.64    & 87.42  & 74.90   \\ \hline
\end{tabular}
}
\caption{Ablation study on the model scale.}
\label{tab:scale}
\end{table*}

\vspace{-2em}
\subsection{Visualizations}

\begin{figure*}[t]
    \centering
    \includegraphics[width=\textwidth]{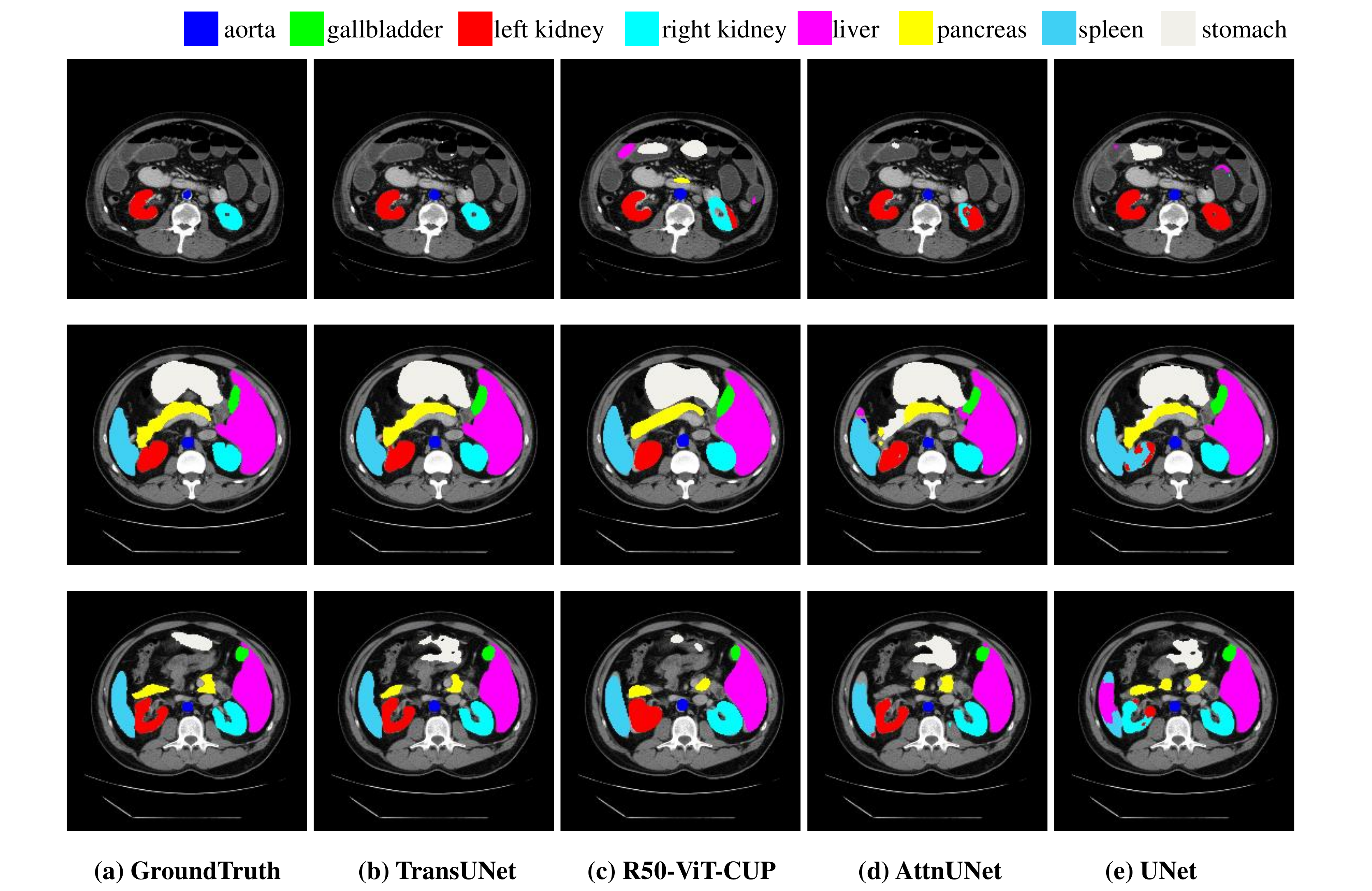}
    \caption{Qualitative  comparison  of  different  approaches by visualization.  From  left  to  right:  (a) Ground Truth, (b) TransUNet, (c) R50-ViT-CUP, (d) R50-AttnUNet, (e) R50-U-Net. Our method predicts less false positive and keep finer information.}
    \label{fig:vis}
\end{figure*}

\begin{table}[t]
\scriptsize
\begin{tabular}{c|c|ccc} 
\hline

Framework         & Average          & RV           & Myo        & LV    \\
\hline
R50-U-Net          & 87.55         & 87.10       & 80.63        & 94.92     \\
R50-AttnUNet      & 86.75         & 87.58       & 79.20       & 93.47       \\
ViT-CUP       & 81.45	& 81.46	& 70.71	 & 92.18    \\
R50-ViT-CUP       &  87.57	& 86.07	& 81.88	& 94.75  \\
TransUNet         & 89.71         & 88.86       & 84.53       & 95.73   \\
\hline
\end{tabular}
\caption{Comparison on the ACDC dataset in DSC (\%).}
\label{tab:acdc}
\end{table}

We provide qualitative comparison results on the Synapse dataset, as shown in Figure \ref{fig:vis}. It can be seen that: 1) pure CNN-based methods U-Net and AttnUNet are more likely to over-segment or under-segment the organs (\emph{e.g.}, in the second row, the spleen is over-segmented by AttnUNet while under-segmented by UNet), which shows that Transformer-based models, \emph{e.g.}, our TransUNet or R50-ViT-CUP have stronger power to encode global contexts and distinguish the semantics. 2) Results in the first row show that our TransUNet predicts fewer false positives compared to others, which suggests that TransUNet would be more advantageous than other methods in suppressing those noisy predictions. 3) For comparison within Transformer-based models, we can observe that the predictions by R50-ViT-CUP tend to be coarser than those by TransUNet regarding the boundary and shape (\emph{e.g.}, predictions of the pancreas in the second row). Moreover, in the third row, TransUNet correctly predicts both left and right kidneys while R50-ViT-CUP erroneously fills the inner hole of left kidney. These observations suggest that TransUNet is capable of finer segmentation and preserving detailed shape information. 
The reason is that TransUNet enjoys the benefits of both high-level global contextual information and low-level details,
while R50-ViT-CUP solely relies on high-level semantic features. 
This again validates our initial intuition of integrating U-Net-like skip-connections into the Transformer design to enable precise localization.

\subsection{Generalization to Other Datasets}

To show the generalization ability of our TransUNet, we further evaluate on other imaging modalities, \emph{i.e.}, an MR dataset ACDC aiming at automated cardiac segmentation. We observe consistent improvements of TransUNet over pure CNN-based methods (R50-UNet and R50-AttnUnet) and other Transformer-based baselines (ViT-CUP and R50-ViT-CUP), which are similar to previous results on the Synapse CT dataset.

\section{Conclusion}
Transformers are known as architectures with strong innate self-attention mechanisms.
In this paper, we present the first study to investigate the usage of Transformers for general medical image segmentation. 
To fully leverage the power of Transformers, TransUNet was proposed, which not only encodes strong global context by treating the image features as sequences but also well utilizes the low-level CNN features via a u-shaped hybrid architectural design.
As an alternative framework to the dominant FCN-based approaches for medical image segmentation, TransUNet achieves superior performances than various competing methods, including CNN-based self-attention methods.

\noindent\textbf{Acknowledgements.} This work was supported by the Lustgarten Foundation for Pancreatic Cancer Research.

%
%
%

{\small
\bibliographystyle{splncs04}
\bibliography{main}}
%

\end{document}